# Optimizing Sensor Placement for Hydrogen Leak Detection in Enclosed Infrastructure: A Comparative Study Using CFD-informed Genetic Algorithm and DeepSets Neural Surrogate

Wang, F. *, Tan Jerome, N., Jordan, T., and Simon, F.
Karlsruhe Institute of Technology, Hermann-von-Helmholtz-Platz 1,
76344 Eggenstein-Leopoldshafen, Germany.

**ABSTRACT**

Hydrogen infrastructure in enclosed environments, such as parking facilities for fuel cell vehicles, presents significant safety challenges due to hydrogen's low ignition energy and wide flammability range. Current monitoring systems are largely reactive, detecting leaks only after hazardous concentrations have formed. This study develops a computational framework for proactive sensor placement optimization by integrating computational fluid dynamics (CFD), genetic algorithm (GA) optimization, and a DeepSets neural surrogate. A CFD database of 180 scenarios was generated for a representative 50 m × 30 m × 3 m garage, covering multiple leak positions, rates (1–150 g/s), and ventilation conditions (ACH = 3–10 $h^{-1}$). Sensor placement was optimized using a multi-objective GA and compared with uniform, random, and surrogate-assisted approaches. The GA achieved a detection rate of 96.1% within 60 s and reduced blind areas to 0.12%, corresponding to an approximately 5% improvement in composite fitness over a uniform baseline. The DeepSets surrogate reproduced near-optimal configurations with a fitness gap below 0.01 while reducing CFD evaluations by 89% and computational time by two orders of magnitude. Detection performance and spatial coverage remained comparable to the GA, demonstrating that surrogate-assisted optimization can retain solution quality while enabling rapid design iteration. Overall, the results show that CFD-informed optimization improves detection effectiveness and reduces sensor requirements compared to conventional layouts. The proposed framework supports scalable deployment and provides a foundation for integrating optimized sensor networks with digital twin systems for real-time monitoring and risk assessment.

## 1 Introduction

The global transition toward sustainable energy has positioned hydrogen as a pivotal energy carrier, with hydrogen fuel cell vehicles experiencing rapid market growth [1]. This expansion necessitates supporting hydrogen energy infrastructure, especially the confined facilities, where hydrogen's unique properties present substantial safety challenges [2]. Hydrogen possesses the lowest ignition energy of any fuel (0.02 mJ, one order of magnitude lower than gasoline), the widest flammability range (4–75 vol.%), and high buoyancy, resulting in rapid ceiling-layer accumulation that can create explosive atmospheres from minimal energy sources [3]. Historical incidents—including the 2019 Gangneung, South Korea explosion [4] and Norway refueling station incident [5]—demonstrate that leaks can rapidly create hazardous conditions in poorly monitored environments.

Current hydrogen safety systems employ reactive detection strategies, triggering alarms only after threshold concentrations (0.4–4.0 vol.%) are exceeded, providing insufficient intervention time (often >60 seconds) for effective emergency response [6][7]. Moreover, conventional sensor placement relies on standard guidelines [8] or empirical grid distributions that would fail to account for facility-specific dispersion patterns, leak source distributions, and ventilation effectiveness, leaving less redundant coverage and critical blind spots, and resulting in suboptimal detection performance.

Recent advances in computational fluid dynamics (CFD) enable high-fidelity simulation of hydrogen dispersion, capturing buoyancy-driven transport, turbulent mixing, ventilation, and geometric constraints [9]. CFD studies have validated concentration predictions within ±10-15% of experimental data for hydrogen releases in confined spaces [10], establishing CFD as a reliable tool for sensor optimization [11][12][13]. Concurrently, evolutionary computation methods — particularly genetic algorithms (GAs) — have demonstrated success in multi-objective sensor network optimization by efficiently exploring large discrete solution spaces [14][15]. Machine learning (ML) based Surrogate-assisted evolutionary computation (SAEC) refers to the integration of surrogate models (also called metamodels) into evolutionary algorithms to solve optimization problems where the evaluation of the objective function is computationally expensive [16]. By approximating the fitness landscape, surrogates significantly reduce the number of costly function evaluations, making it feasible to apply evolutionary algorithms to real-world problems involving simulations (e.g., CFD, finite element analysis) or time-consuming experiments. DeepSets neural network architecture can be, and has been, used as a surrogate model in the context of SAEC, though its role is specific and distinct in formulation and inductive bias that approximate the objective function itself [17][18]. However, comprehensive studies integrating validated CFD databases with multi-method optimization approaches (combining evolutionary and machine learning techniques) remain limited for hydrogen safety applications.

Research gaps persist in the literature: (1) Most studies evaluate configurations against a limited set (typically dozens at most) of CFD scenarios [11][12][13], insufficient to capture variability in leak positions, rates, and ventilation conditions typical of hydrogen facilities; (2) Previous studies predominantly employ single-objective formulations, optimizing either detection rate or simplified efficiency metrics. Temporal performance—critical for early warning—is often neglected or treated in an aggregated manner, whereas the present work adopts a multi-objective formulation that explicitly balances detection, timing, and coverage [13][15]; (3) Systematic, controlled comparisons of sensor placement optimization methods (GA, DeepSets, baselines) under identical conditions remain limited in literature.

This study addresses these gaps by developing a comprehensive framework for sensor placement optimization in a representative underground parking garage. Specific objectives are: (1) Generate a 180-scenario CFD database spanning leak rates, positions (12 locations: corners, walls, mid-rows), and ventilation conditions; (2) Implement and compare genetic algorithm (GA) with multi-objective fitness (detection probability, spatial coverage, temporal response) against DeepSets Neural Surrogate and heuristic baseline methods (random, uniform); (3) Evaluate performance across detection rate, timing, and coverage metrics.

The remainder of this paper is organized as follows: Section 2 describes CFD modeling, optimization algorithms, and evaluation metrics. Section 3 presents CFD database characteristics, optimization results, and performance comparisons. Section 4 discusses interpretation and implications. Section 5 with key findings and recommendations for hydrogen infrastructure safety.

# 2 Methodology

The methodology employs a three-stage approach to optimize sensor placement for hydrogen leak detection in an enclosed facility: (1) CFD simulations to generate a comprehensive database of leak scenarios, (2) sensor optimization using GA and DeepSets alongside baseline methods, and (3) comparative performance evaluation using defined metrics for detection rate, timing, and coverage.

## 2.1 CFD modelling

### 2.1.1 Facility Configuration

The modeled facility is an underground parking garage with dimensions 50 m (length) × 30 m (width) × 3 m (height), representative of typical enclosed hydrogen infrastructure housing fuel cell vehicles. The parking garage accommodates approximately 60 vehicles (Figure 1). The layout includes two horizontal traffic lanes, and two vertical connecting lanes. Structural columns (0.5 m × 0.5 m cross-section) are positioned on a 6 m × 6 m grid to provide ceiling support, creating potential flow obstructions and sensor placement constraints.

Mechanical ventilation is supplied through a corner across the facility, with exhaust vents located at diagonal corner to promote crossflow. Three ventilation rates were modeled: ACH = 3 $h^{-1}$ (minimal code-compliant ventilation), ACH = 6 $h^{-1}$ (typical operation during accidents), and ACH = 10 $h^{-1}$ (emergency mode). This configuration is representative of European [19] and North American parking garage standards [20]. Six jet fans with 10 m/s flow speed are installed under the ceiling and provide the momentum of gas to enhance the circulation.

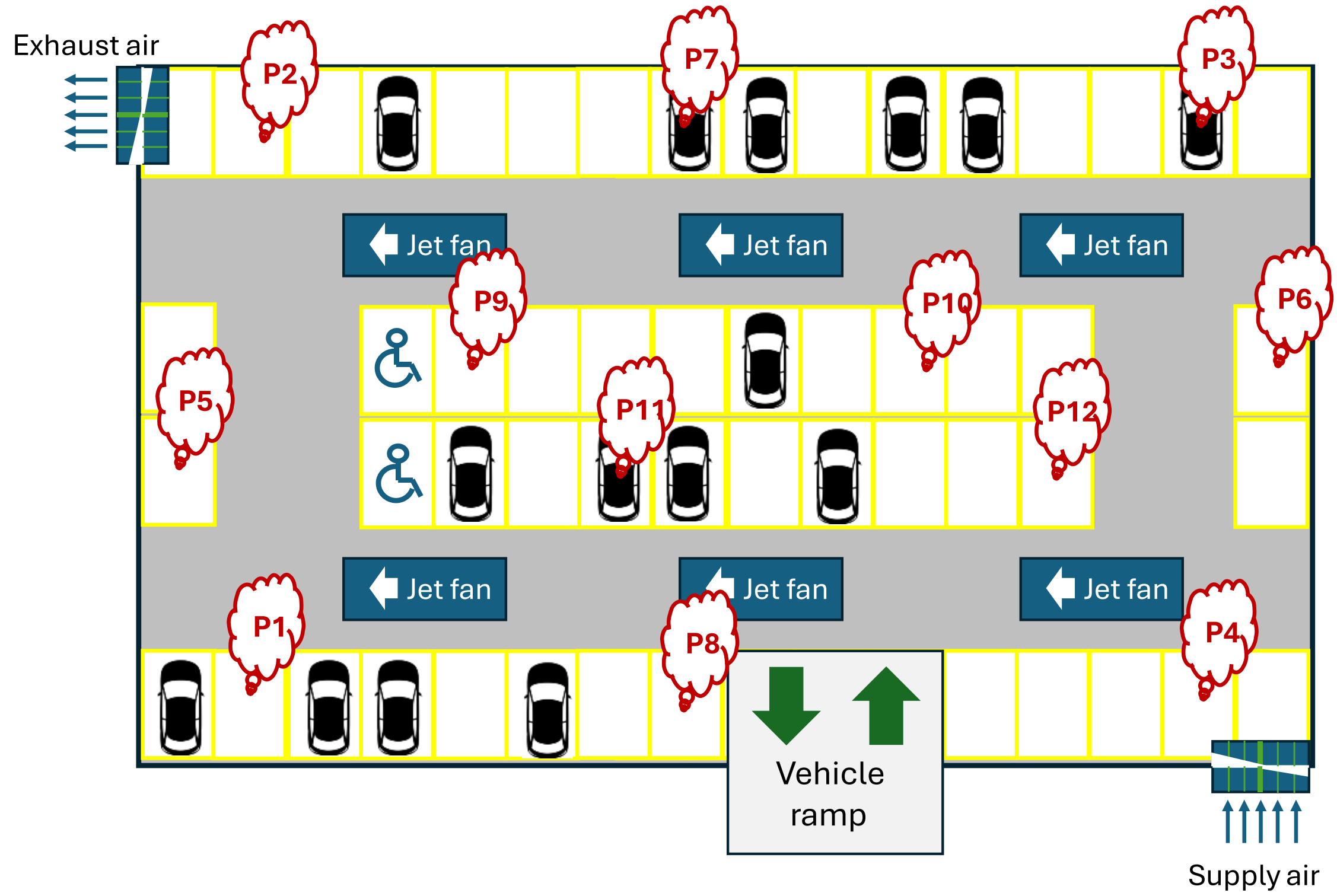

Figure 1 Underground parking garage geometry and dimensions, showing parking layout, ventilation configuration, and hypothetic leak positions.

### 2.1.2 Governing Equations

The simulations solve the three-dimensional, time-dependent compressible Navier-Stokes equations for mass (Equation 1), momentum (Equation 2), and species conservation (Equation 3). The species transport equation models hydrogen diffusion and advection, with source terms representing leak mass flows.

$$\frac{\partial \rho}{\partial t} + \nabla \cdot (\rho \mathbf{u}) = 0 \qquad \text{Equation 1}$$

$$\frac{\partial \rho \mathbf{u}}{\partial t} + \nabla \cdot (\rho \mathbf{u}\mathbf{u}) = -\nabla p + \nabla \cdot \boldsymbol{\tau} + \rho \mathbf{g} \qquad \text{Equation 2}$$

$$\frac{\partial \rho Y_i}{\partial t} + \nabla \cdot (\rho \mathbf{u} Y_i) = \nabla \cdot (\rho D_i \nabla Y_i) \qquad \text{Equation 3}$$

where ρ is density, **u** is velocity vector, p is pressure, **τ** is stress tensor, **g** is gravitational acceleration, $Y_i$ is species mass fraction, and Di is diffusion coefficient. The standard k-ε turbulence model was employed due to its validated accuracy for buoyancy-driven flows in confined spaces [9], with transport equations for turbulent kinetic energy (k) and dissipation rate (ε) solved alongside momentum and species equations.

Buoyancy effects were incorporated using the Boussinesq approximation, critical for hydrogen dispersion due to its low density (0.082 kg/m³ at 298 K, 14× lighter than air). The ideal gas law governed density variations with local hydrogen concentration, enabling accurate representation of buoyancy-driven stratification and ceiling accumulation.

The used CFD tool GASFLOW was validated against published hydrogen release experiments [10], involving controlled leaks in enclosed spaces. Simulations replicated experimental geometries and leak rates, achieving concentration predictions within ±10%. Error analysis confirmed the model's accuracy for buoyancy-driven flows. This validation confirms the model's suitability for sensor optimization applications.

### 2.1.3 Mesh, Boundary Conditions and Scenario Matrix

The computational domain encompasses the full garage geometry, discretized using an orthogonal structured mesh with baseline resolution 25 cm to capture buoyancy-driven flows. Mesh independence was verified by comparing peak concentrations across three mesh resolutions: 36k cells (coarse), 288k cells (medium), and 972k cells (fine). The medium mesh yielded concentration differences <5% compared to the fine mesh (Table 1, in the case of ACH=6 $h^{-1}$, leak mass flow 50 g/s, leak position 10), with computational time of ~1.5 hours per simulation of 1-minute physical problem on a Server (Intel Xeon, 32 GB RAM, 1 node 16 processors 3.2 GHz), balancing accuracy and efficiency.

Ventilation exhaust is specified as mass flow outlets with uniform flow distribution matching target ACH rates; fresh air supply set to mass flow inlets but 90% mass flow rate of exhaust, to keep the minus pressure of the garage. The vehicle ramp is modeled as a pressure condition 1 atm, which can supplement the mass loss from exhaust. Initial conditions assume quiescent air at 293 K and 1 atm. No-slip and adiabatic conditions are applied at walls, with standard wall functions for turbulence. Gravitational acceleration was set to 9.8 $m/s^2$ in the negative Z-direction.

Table 1 Mesh Independence Study

| Sort | Mesh size (cells) | peak concentration at t = 60 s | CPU-Time | Relative Error |
|---|---|---|---|---|
| Coarse | 100 X 60 X 6 (36K) | 13.1% | 0.25 h | 58.3% |
| Medium | 200 X 120 X 12 (288K) | 8.5% | 1.5 h | 3.6% |
| Fine | 300 X 180 X 18 (972K) | 8.2% | 13 h | Baseline |

A total of 180 scenarios (180 = 12 X 5 X 3) were generated by full factorial combination of three parameters: 12 leak positions, 5 leak rates (1, 30, 50, 100, 150 g/s), and 3 ventilation levels (ACH = 3, 6, 10 $h^{-1}$). The 12 leak positions (as seen in Figure 1) were strategically distributed to represent realistic vehicle locations: four corner positions (P1-P4), four wall-adjacent positions (P5-P8), and four mid-row parking position (P9-P12). This distribution inherently emphasizes central parking areas where multiple leak sources overlap, reflecting actual vehicle density patterns without requiring explicit probability weighting. All leaks were positioned at Z = 0.5 m (fuel tank height) to simulate realistic failure modes, and all 12 leak positions used for all 5 leak rates and all 3 ACH levels.

The calculation was run around 10 minutes before leak to get the initial filed under minimal code-compliant ventilation ACH = 3 $h^{-1}$. After $H_2$ leak, each simulation was run for 60 seconds with concentration fields saved at 1-second intervals (60 timesteps per scenario). Outputting 3D concentration fields C (x, y, z, t) stored in HDF5 format for optimization input.

## 2.2 Sensor Optimization Methods

### 2.2.1 Baseline Methods

Two baseline configurations provide benchmarks for optimization method evaluation:

**Random Placement**: 15 sensors positioned randomly within feasible regions (avoiding columns, maintaining minimum 5 m spacing) using Monte Carlo sampling.

**Uniform Grid**: 15 sensors arranged in a regular 5 × 3 grid pattern at ceiling level, representing conventional engineering practice. Spacing of 10 m (X-direction) × 10 m (Y-direction) ensures uniform coverage.

Both baselines enforce constraints: ceiling mounting (Z = 2.75 m), minimum inter-sensor spacing (5 m), and avoidance of structural columns (>0.5 m clearance).

### 2.2.2 Genetic Algorithm Optimization

The GA configuration employs real-valued encoding where each individual represents a sensor configuration as a 15×3 array specifying (X, Y, Z) coordinates for 15 sensors. Population size was set to 150 individuals, providing sufficient diversity while maintaining computational tractability. Evolution proceeds for 100 generations with operators: tournament selection (size = 3), uniform crossover (probability = 0.7), and Gaussian mutation (probability = 0.15). These parameters were determined through preliminary sensitivity analysis.

The multi-objective fitness function (Equation 4) balances four performance criteria:

$$F = \alpha \cdot P_{detection} + \beta \cdot C_{coverage} + \gamma \cdot S_{timing} - \Phi_{penalty} \qquad \text{Equation 4}$$

where empirically tuned weights α = 0.35, β = 0.30, γ = 0.35 sum to unity, and components are normalized to [0, 1]. In this study, the weighting factors were selected empirically to assign approximately equal importance to each objective, as no prior information was available to justify prioritizing one objective over the others. This choice provides a balanced optimization framework in which improvements in each objective contribute comparably to the overall fitness. It should be noted that the weighting factors are application-dependent and can be adjusted to reflect specific design priorities. For example, if early detection is considered more critical than the other objectives, a larger weight can be assigned to the corresponding objective. The influence of the selected weights on the optimization results can be assessed through a sensitivity analysis in section 3.2.3, which quantifies the robustness of the obtained solutions and identifies the relative importance of each objective.

**Detection Probability ($P_{detection}$):** Fraction of scenarios where at least one sensor exceeds the detection threshold (0.1 vol.%):

$$P_{detection} = \frac{1}{N}\sum_{i=1}^{N} 1\left(\max\left(C_{i,sensors}\right) > \ 0.001\right) \qquad \text{Equation 5}$$

where unit 1 is the indicator function.

**Coverage Score ($C_{coverage}$)**: Evaluates spatial coverage by assessing sensor redundancy in critical parking zones. Each zone is sampled at a certain number of points (e.g. 50 × 30 grid), and the number of sensors within detection radius (CFD results based above detection limit 0.1 vol.% $H_2$ cloud at the ceiling, ~7.85 m) of each point is counted. The score rewards optimal redundancy (2-3 sensors) while penalizing insufficient coverage (0 sensor) or excessive redundancy (>3 sensors):

$$C_{coverage} = \frac{1}{N_{zones}}\sum_{z=1}^{N_{zones}} \frac{1}{N_{points}}\sum_{p=1}^{N_{points}} f\left(n_{sensors}(p,z)\right) \qquad \text{Equation 6}$$

where $N_{zones}$ = critical zones (CFD results based low flammability limit 4 vol.% $H_2$ cloud at the ceiling, here 1 zone centered at parking area (25,15) with dimensions 49.8×29.8 m), $N_{points}$ = 50 ×30 sample points per zone, $n_{sensors}(p,z)$ is the count of sensors within 7.85 m radius of point p in zone z, and the scoring function f is defined as:

$$f(n) = \begin{cases} 0.0 & if\ n = 0 \text{ (blind)} \\ 0.8 & if\ n = 1 \text{ (sufficient coverage)} \\ 1.0 & if\ 2 \le n \le 3 \text{ (optimal redundancy)} \\ 0.5 & if\ n > 3 \text{ (excessive redundancy)} \end{cases} \qquad \text{Equation 7}$$

This formulation prioritizes adequate redundancy (2-3 sensors per location) to ensure robustness against single-sensor failures while avoiding over-instrumentation.

**Timing Score ($S_{timing}$)**: Rewards early detection by evaluating transient concentration data across all 180 scenarios. For each scenario, the detection time $t_i$ is determined as the first timestep where any sensor exceeds the threshold (0.1 vol.%). Scenarios where no sensor triggers within the simulation period receive a penalty time equal to the maximum timesteps ($t_{max}$ = 60, equivalent to 60 seconds):

$$t_i = \begin{cases} \min\limits_{t\in[0,t_{max}]} \left\{t\colon \left(\max\limits_{j}\left(C_{i,j}\right) > \ 0.001\right)\right\} & \text{if detection occurs} \\ t_{max} & \text{otherwise (non} - \text{detection penalty)} \end{cases} \qquad \text{Equation 8}$$

where $C_{i,j}(t)$ is the concentration at sensor j during scenario i at timestep t. The average detection time is computed across all scenarios, then normalized using a reciprocal transformation that ensures faster detection yields higher scores:

$$\mathrm{S_{timing}} = e^{-3\frac{\bar{t}_{detect}}{t_{max}}} \quad \text{Equation 9}$$

where $\bar{t}_{detect} = \left(\frac{1}{\mathrm{N}}\right)\Sigma t_i$ is the mean detection timestep. The time constant 3 is empirical. This normalization maps: instant detection ($\bar{t}_{detect} = 0$) → score=1.0, median time ($\bar{t}_{detect} = t_{max}/2$) → score=0.223, and maximum time ($\bar{t}_{detect} = t_{max}$) → score=0.05, providing non-linear emphasis on early warning performance.

**Penalty Term ($\mathbf{\Phi_{penalty}}$)**: Enforces hard constraints by penalizing: spacing violations (sensors closer than 5 m), infeasible positions (inside columns or outside facility bounds), coverage violations (primarily equals the blind-area ratio, augmented by a linear penalty when double+ coverage falls below 80%), and wall proximity violations (the minimum distance between sensor and wall equals 1 m). These components are combined via weighted summation with weights $w_{\text{spacing}} = 0.2$, $w_{\text{feasibility}} = 0.2$, $w_{\text{coverage}} = 0.3$, and $w_{\text{wall}} = 0.3$ ensuring the total penalty remains within [0, 1], where 0 denotes a fully feasible, high-coverage configuration and 1 represents the worst possible violation scenario.

**Constraint Handling:** Sensor positions are initialized within feasible regions and repaired whenever mutation produces infeasible solutions (e.g., outside the domain or within obstacles). During crossover, infeasible offspring are corrected by inheriting feasible coordinates from parent solutions. This combined repair–penalty strategy preserves feasibility across the population while still permitting exploration near constraint boundaries, thereby improving convergence in constrained optimization problems.

### 2.2.3 DeepSets Neural Surrogate

- **DeepSets Architecture**

A key constraint in sensor placement optimization is permutation invariance: the fitness of a configuration must not depend on the arbitrary order in which sensors are listed. A conventional feedforward network applied to a concatenated sensor vector would violate this property, producing different predictions for configurations that differ only in sensor ordering.

To enforce permutation invariance, we adopt the DeepSets architecture [18]. The model applies a shared encoder network φ independently and identically to each sensor position, then aggregates the resulting representations via mean pooling before passing the pooled vector through a decoder network ρ:

$$f(\mathrm{S}) = \rho\big(\mathrm{mean}\{\mathrm{s} \in \mathrm{S}\}\varphi(\mathrm{S})\big) \quad \text{Equation 10}$$

where $S = \{s_1, \ldots, s_N\}$ is the set of N sensor positions, each $s_i \in \mathbb{R}^3$ encoding (x, y, z) coordinates. The encoder φ consists of two linear layers with layer normalization and ReLU activation (dimensions 3 → 64 → 128), and the decoder ρ applies three linear layers (128 → 128 → 64 → 4) with a final sigmoid activation. The four decoder outputs correspond to detection rate, coverage score, speed score, and predicted constraint penalty respectively. The composite fitness is computed as a weighted sum with weights (0.35, 0.30, 0.35, −1.0).

The entire network is implemented in pure NumPy without deep learning frameworks, which simplifies deployment and avoids hardware-specific dependencies. Training uses the Adam optimizer with a weighted mean-squared error loss that assigns 2× weight to samples with above-median speed scores, compensating for the relative scarcity of high-speed detection configurations in the training data.

- **Training Data Construction**

Training data are derived from the GA population history rather than from dedicated CFD runs. The GA accumulates many evaluated configurations across its generations; these represent a biased but informative sample of the design space, concentrated in regions the evolutionary search found promising. This bias is intentional by design — the surrogate is not required to emulate the full design space, but only to guide local gradient-based refinement from where the GA left off; the trust-region constraint (described in the Trust-Region Optimization subsection below) ensures the optimiser does not venture beyond the well-sampled region.

From this history, a speed-biased stratified subsample of 800 configurations is selected for re-evaluation on the true CFD simulator. The sampling strategy draws 30% of examples from the top-fitness GA configurations (which tend to exhibit high speed scores) and the remaining 70% via fitness-stratified sampling across the full fitness range. This dual strategy ensures the surrogate receives both high-quality examples at the peak of the fitness landscape and diverse coverage across the broader distribution, improving generalization. The labelled dataset is split 85/15 into training and validation sets, and training proceeds for up to 500 epochs with early stopping (patience = 60 epochs) on validation loss.

- **Trust-Region Optimization, Restart Augmentation, and Coverage Bonus**

Sensor placement is framed as a continuous optimization problem solved via L-BFGS-B with finite-difference gradients through the surrogate. A quadratic trust-region penalty ($\lambda = 0.05$, $r = 5$ m) confines each restart to the neighborhood of a known-good seed configuration, preventing the surrogate overestimation that produced a −0.11 fitness gap in unregularized preliminary experiments. Wall-margin constraints are enforced through hard optimizer bounds, a penalty gradient, and anchor clipping. Of the 20 restarts, 15 are seeded from top-fitness GA configurations and 5 from uniformly random positions to ensure coverage of facility regions underrepresented in the GA history. A spatial coverage grid bonus (25×15 cells, $w_{cov} = 2.5$) adds a Gaussian proximity kernel term to the objective, providing targeted gradients that pull sensors toward blind spots. Together these measures reduce the surrogate-to-CFD fitness gap to $\Delta = +0.017$ and the facility blind spot to 0.07%. Full implementation details are provided in the Supplementary Material.

## 2.3 Performance Evaluation Metrics

Sensor configurations are assessed across three categories, and all metrics are computed for each method (Random, Uniform, GA, DeepSets):

**Detection Performance**: Detection rate (percentage of 180 scenarios exceeding threshold), false negative rate (undetected leaks), and detection rate stratified by leak rate.

**Timing Performance**: Mean and median detection time, best/worst detection time, time-to-95% cumulative detection, and early warning gain (seconds saved compared to uniform baseline).

**Coverage Metrics**: Total coverage (floor area within detection radius of any sensor), redundant coverage (areas covered by 2+ sensors), blind spot area (uncovered regions), and coverage histogram showing single/double/triple+ coverage distribution.

# 3 Results

## 3.1 CFD Simulation Results

### 3.1.1 Hydrogen Dispersion Evolution

The CFD simulations captured the characteristic buoyancy-driven dispersion of hydrogen in the enclosed parking garage across all 180 scenarios. Figure 2 illustrates typical hydrogen dispersion patterns on top surface ($Z = 2.75$ m) where sensors are mounted at selected time intervals. Upon release from the vehicle fuel tank height ($Z = 0.5$ m), hydrogen rapidly ascended toward the ceiling due to its release momentum and low-density resulting buoyancy force. The vertical rise phase occurred mostly within several seconds depending on leak rate, with hydrogen reaching the ceiling and forming a stratified layer.

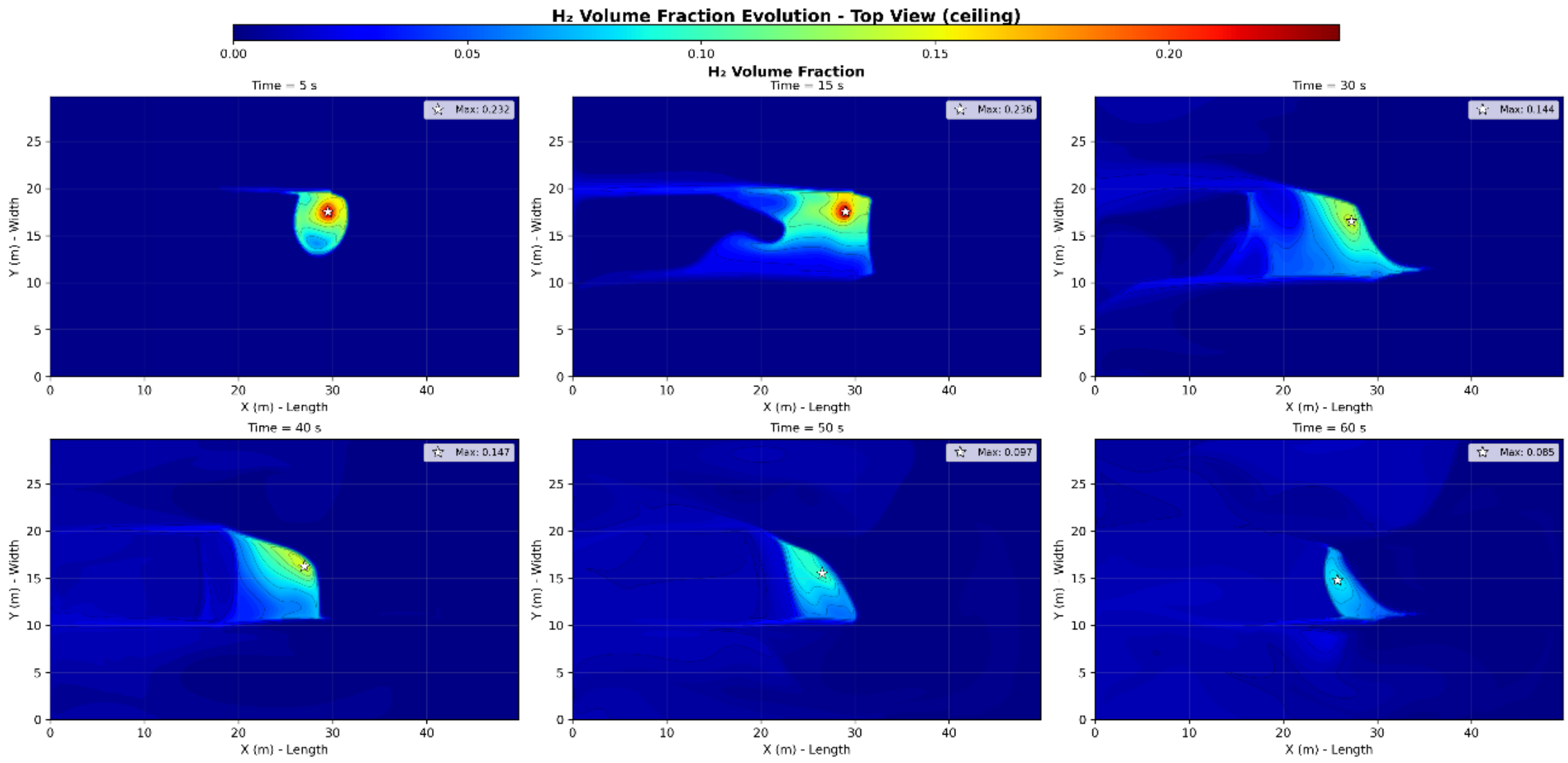

Figure 2 Hydrogen volume fraction under ceiling over time of the case ACH=6 $h^{-1}$, leak flow 50 g/s and leak position 10.

Lateral spreading along the ceiling plane exhibited complex behavior influenced by jet fan, ventilation flow patterns and structural obstacles. The ceiling layer extended radially from the leak point, with the flow circulation that mixed hydrogen in the whole region. Under moderate ventilation (ACH = 6 $h^{-1}$), hydrogen accumulated persistently in ceiling with peak concentrations ranging from 8.5–23.6 vol.%, significantly exceeding the lower flammability limit (LFL = 4 vol.%). The jet fan-assisted ventilation system promoted crossflow patterns that enhanced mixing and directed hydrogen-rich air toward exhaust vents, though effectiveness varied spatially in regions distant from jet fan coverage.

The CFD database provides spatially and temporally resolved concentration fields C (x, y, z, t) that serve as ground truth for evaluating sensor configurations, enabling assessment of detection probability, timing, and coverage across the full range of realistic leak scenarios.

### 3.1.2 Effect of Parameters

Parametric analysis quantified the influences of leak position, ventilation rate, and leak rate on hydrogen dispersion dynamics, as illustrated in Figure 3.

**Effect of Leak Position** (Figure 3 upper): Spatial location of the leak source significantly modulated dispersion patterns and flammable cloud geometry. Corner positions (P3) generated around 100% higher peak concentrations compared to central positions (P11) due to confinement by perpendicular walls that restricted lateral dispersion and promoted vertical stratification. Wall-adjacent positions (P6) exhibited intermediate behavior, with concentrations about 19% above central leaks.

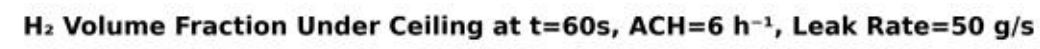

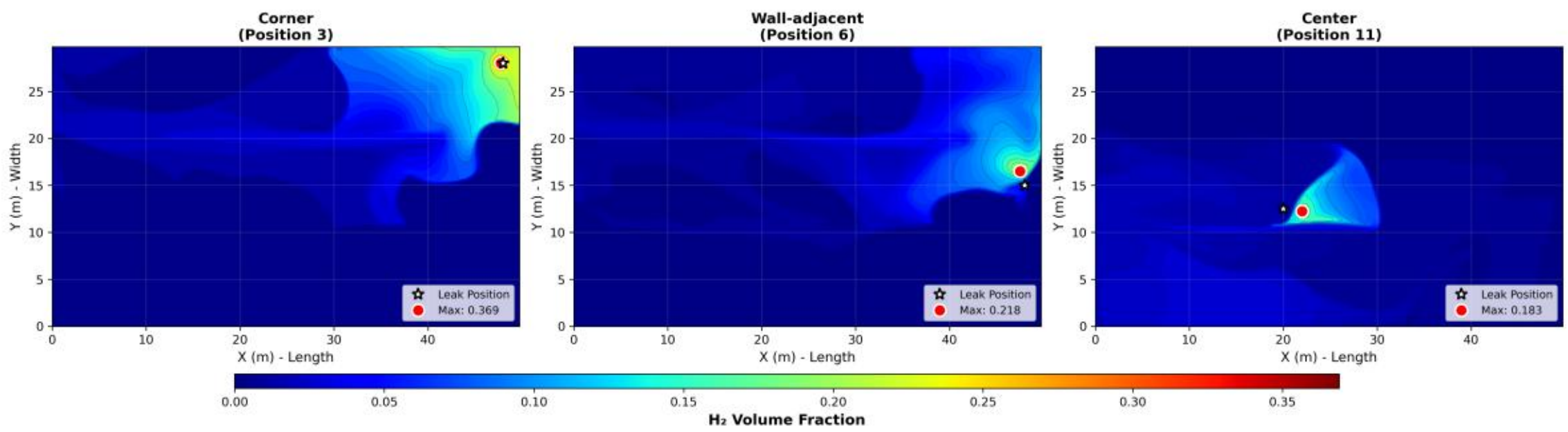

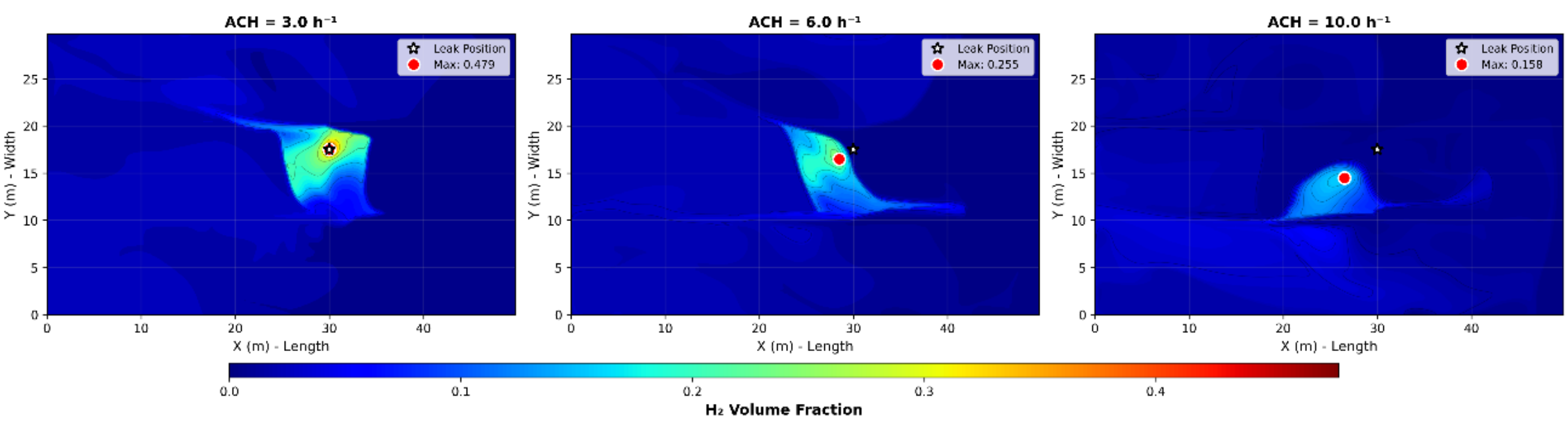

Figure 3 Parameter effect on hydrogen volume fraction under ceiling (Z=2.75)

**Effect of Ventilation Rate** (Figure 3 lower): Mechanical ventilation rate, parameterized by ACH, exerted dominant control over concentrations and flammable cloud persistence. Increasing ventilation from ACH = 3, 6, 10 $h^{-1}$ reduced peak concentrations by 73% (from 47.9 vol.% to 15.8 vol.% in case of leak rate = 100 g/s). The dilution effect followed approximately linear scaling with ventilation rate. However, ventilation effectiveness was spatially non-uniform, with blind zones in corners exhibiting higher concentrations than well-ventilated regions. Jet fan placement created preferential flow paths that directed hydrogen-rich air toward exhaust vents, mixing the concentrations along these corridors.

**Effect of Leak Rate (**Figure 2 right bottom and Figure 3 lower middle**)**: Leak mass flow rate exerted strong influence on peak concentration magnitude and dispersion kinetics. Increasing leak rate from 50 g/s to 100 g/s elevated peak concentrations from 8.5 vol.% to 25.5 vol.% (3× increase), though the relationship was sub-linear due to enhanced mixing at higher injection velocities. Time to reach steady-state decreased, indicating faster equilibration driven by higher momentum flux and stronger buoyancy-induced convection. Detection time—critical for early warning systems—decreased across all positions and ventilation rates, demonstrating that higher leak rates provide shorter intervention windows but are easier to detect. Notably, leak rates ≥50 g/s produced flammable volume (>4 vol.%) even under emergency ventilation (ACH = 10 $h^{-1}$), highlighting the necessity of rapid detection and automated response systems.

## 3.2 Sensor Optimization Results

This section demonstrates how GA and DeepSets exploit the CFD-derived spatial patterns to achieve superior sensor placements compared to heuristic baselines.

### 3.2.1 Genetic Algorithm Evolution

Figure 4 illustrates the convergence behavior of the genetic algorithm over 100 generations, tracking three key performance indicators:

- **Average fitness trajectory (blue line)**: Shows the mean fitness across the entire population of each generation. This curve demonstrates steady improvement from an initial average fitness of 0.552 in Generation 0 to 0.826 by Generation 100, representing a 49.6% improvement. The trajectory exhibits characteristic GA behavior with rapid initial gains (Generations 0-30) as the algorithm discovers promising regions of the search space, followed by gradual refinement (Generations 30-100) as solutions converge toward local optima.
- **Best fitness trajectory (green line)**: Tracks the highest-performing individual in each generation, rising from 0.675 (Generation 0) to a final optimal value of 0.829 with an improvement 22.8%. The curve shows a stepwise improvement pattern typical of evolutionary algorithms, with plateaus indicating periods of exploitation (refining existing good solutions) punctuated by jumps representing beneficial mutations or crossover events that discover superior configurations. The best solution remained almost stable through the final generations, indicating convergence.
- **Analysis of convergence metrics. Fitness range compression**: The gap between best and worst individuals narrowed from 0.125 to 0.005, indicating population convergence around high-quality solutions while maintaining sufficient variation for robustness. **Convergence rate**: Calculated as the average fitness improvement per generation over the final 40 generations (0.001/generation), indicating stable refinement rather than oscillation. **Stagnation behavior**: The best solution almost remained unchanged for 15 consecutive generations before termination, satisfying the convergence criterion (almost no improvement for >10 generations) while avoiding premature termination.

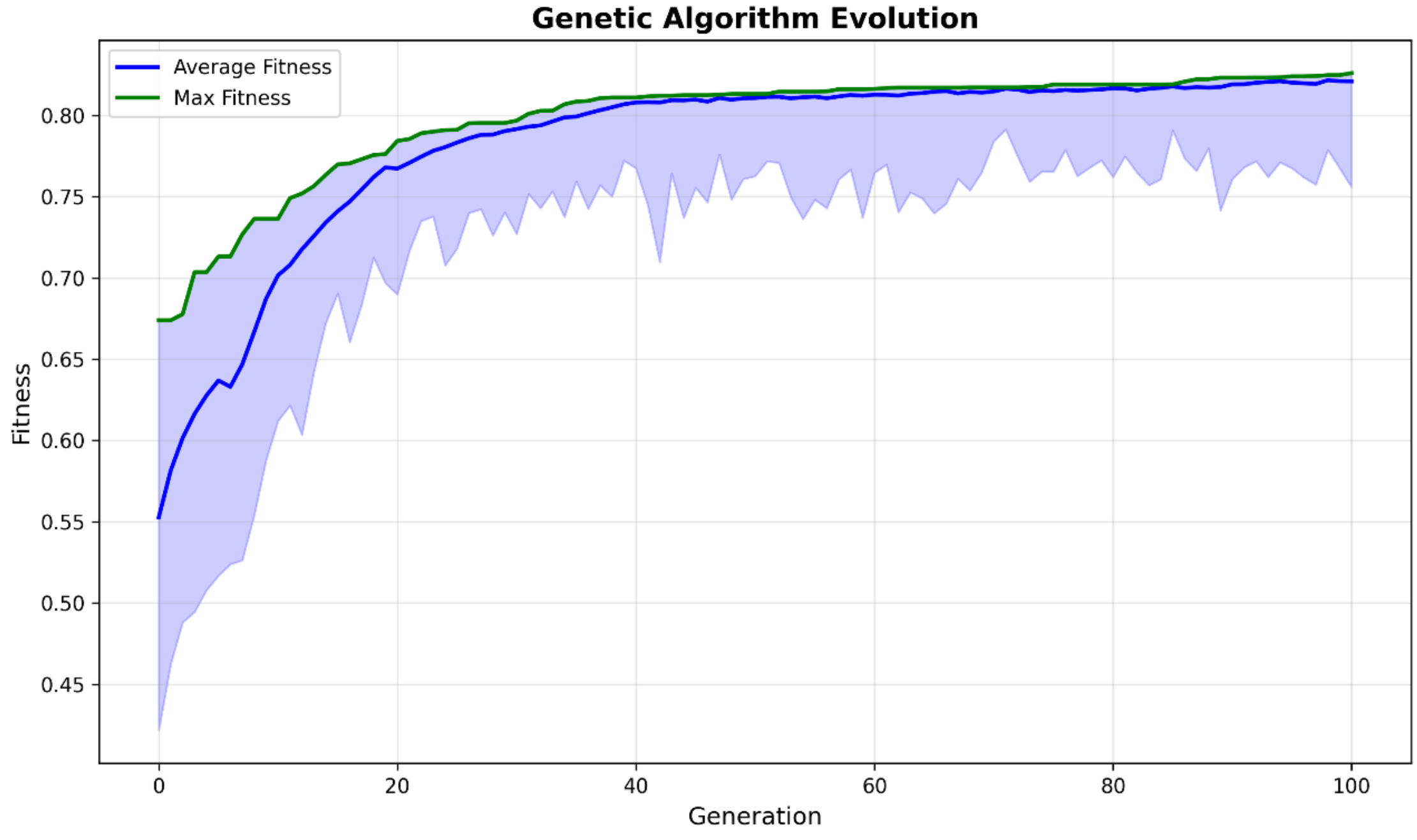

Figure 4 GA fitness evolution over generations

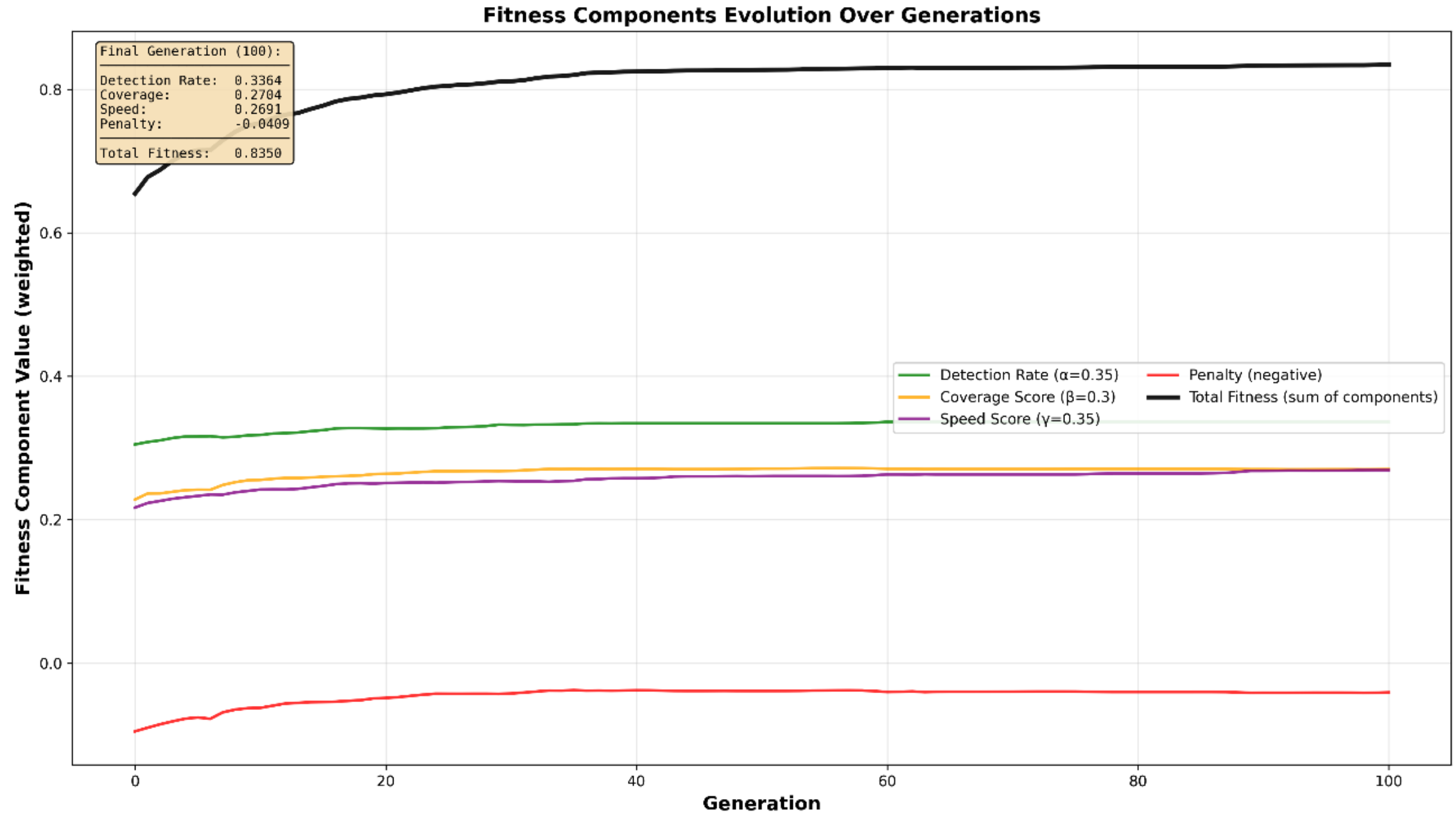

Figure 5 GA fitness components over generations

These statistics confirm that the GA achieved effective optimization within the allocated computational budget (150 individuals × 100 generations = 15,000 fitness evaluations), with the population transitioning from random exploration to focused exploitation of high-performing sensor placement strategies.

Figure 5 presents the evolution of GA fitness components over 100 generations. Total fitness is driven primarily by detection probability (green) improvements. The spatial coverage (orange) and timing score (purple) also exhibit steady improvement, though each contribution remains approximately 2/3 that of detection rate component. Notably, the penalty term (red) decreases significantly at beginning, contributing substantially to final fitness improvement. Convergence stabilizes at the end, confirming the adequacy of the algorithmic termination criteria.

Table 2 Fitness score comparison across sensor placement methods

| Configuration | Fitness Score | Relative to Best |
|---|---|---|
| Random Placement | 0.5724 | 69.31% |
| Uniform Grid | 0.7822 | 94.71% |
| GA Optimized | 0.826 | 100.0% |

The multi-objective fitness function quantitatively demonstrates the superiority of optimization-based methods, as shown in Table 2. GA optimization achieves 30% improvement over random baseline and 5.3% over conventional uniform placement.

### 3.2.2 DeepSets Training

Figure 6 shows the training dynamics of the DeepSets surrogate. Training converged in approximately 290 epochs, with early stopping triggered after 60 consecutive epochs without validation loss improvement, well before the 500-epoch budget was exhausted. The left panel displays the weighted Mean Squared Error (MSE) loss for both the training and validation sets. Both curves drop sharply within the first 20–30 epochs as the network rapidly learns the dominant structure of the fitness landscape, before plateauing near 0.001 from around epoch 50 onward. The close tracking between training and validation loss throughout indicates that the model generalises well and does not overfit, despite the relatively compact training set of 800 labelled configurations.

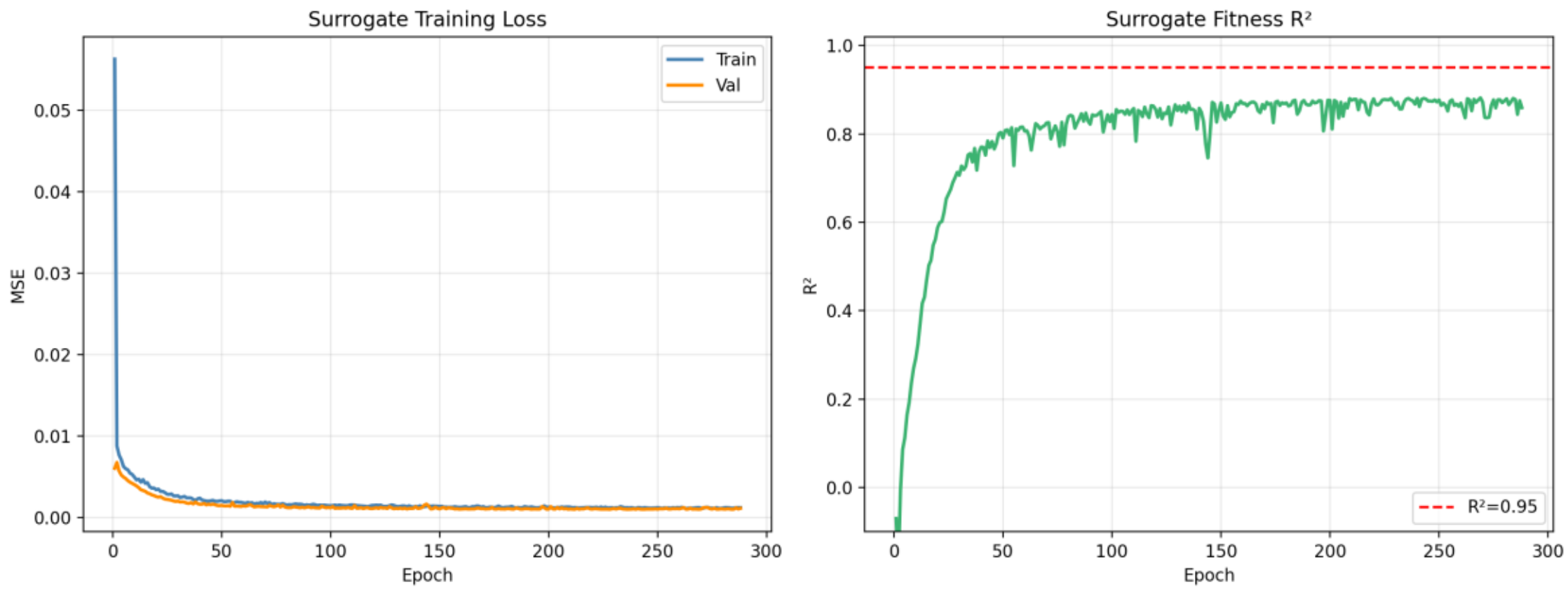

Figure 6 DeepSets surrogate training loss and fitness $R^2$

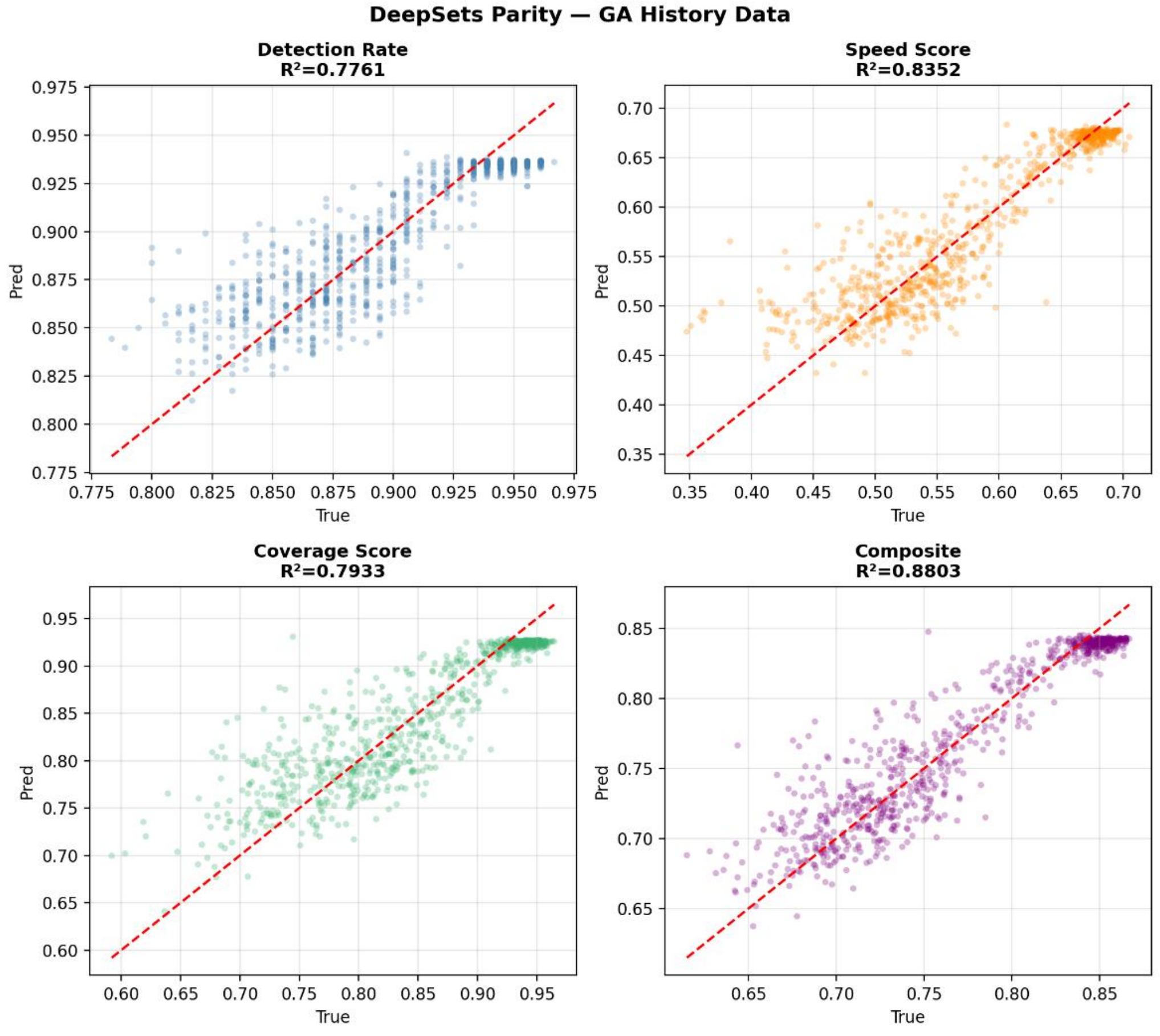

Figure 7 Component parity plots

The right panel in Figure 6 shows the corresponding validation $R^2$, which rises rapidly from near zero to approximately 0.70 within the first 30–40 epochs, then stabilises in the range 0.83–0.87 for the remainder of training. The component-level parity plots ( Figure 7) reveal $R^2$ values of 0.78 for detection rate, 0.84 for speed score, 0.79 for coverage score, and 0.88 for the composite fitness. Although these fall below the target threshold of 0.95 (dashed red line), indicating that the surrogate does not fully capture the variance of the true fitness function, this is expected given that the GA population converges toward a narrow band of high-performing configurations, reducing the effective diversity of the training data. The trust-region constraint ensures that the gradient optimiser operates only within well-sampled regions where surrogate accuracy is highest, making the achieved $R^2$ sufficient for effective gradient-based search. The final DeepSets-predicted configuration is always re-evaluated against the true CFD fitness function, so surrogate error does not propagate to the reported results.

### 3.2.3 Weight Sensitivity Analysis

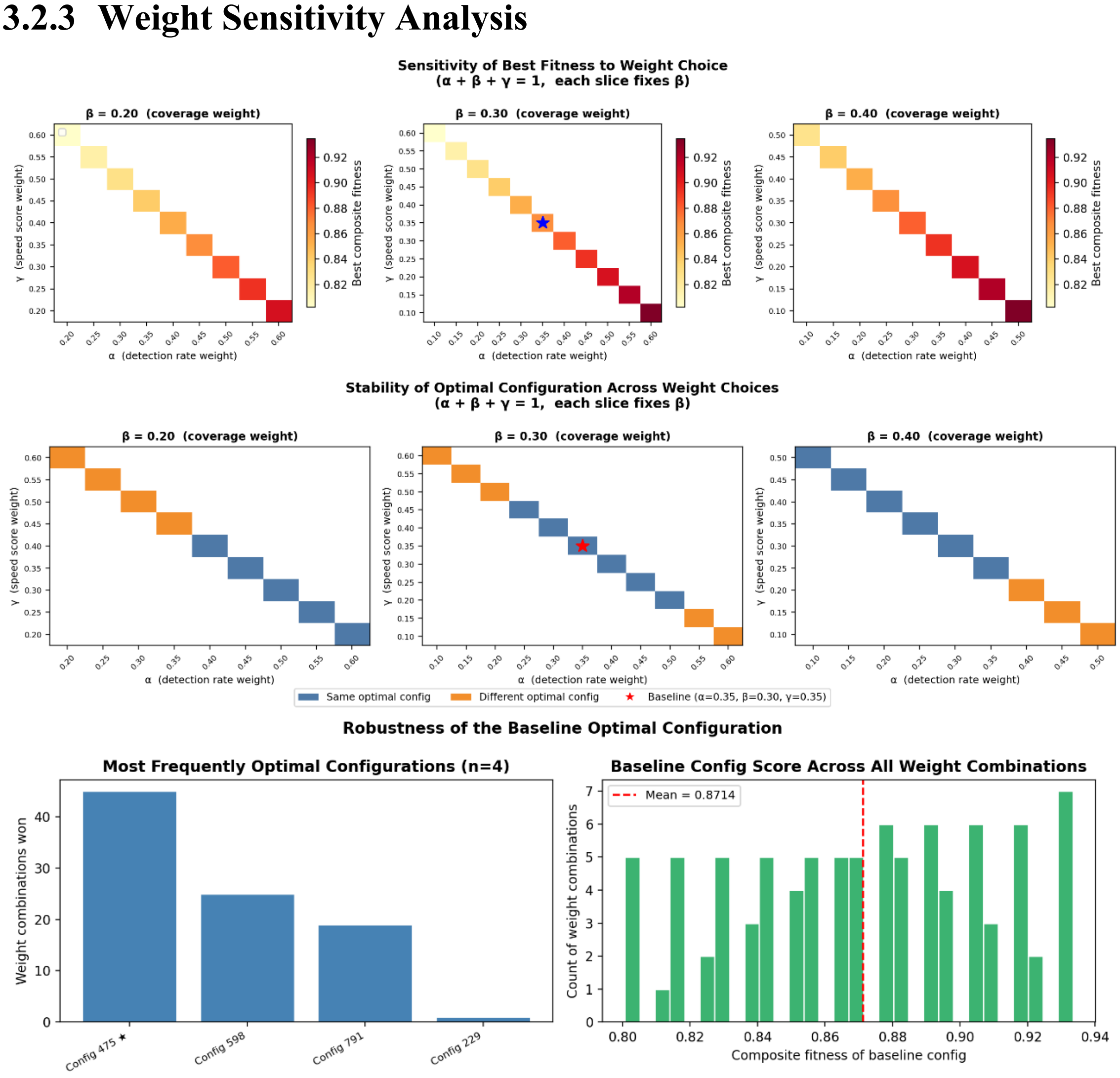

Figure 8 Weight sensitivity analysis results

To assess whether the chosen weights in the fitness function constitute a critical hyperparameter, the three component scores from all 800 evaluated configurations were re-ranked across 90 weight combinations ($\alpha, \beta, \gamma \in \{0.10, 0.15, \ldots, 0.60\}$, $\alpha + \beta + \gamma = 1$).

Figure 8a (top) hows the best achievable composite fitness across the α, β, γ weight grid for three fixed values of β (coverage weight: 0.20, 0.30, 0.40). The achievable fitness varies between 0.80 and 0.93 across the weight space, peaking when both detection rate and speed score receive higher weights — consistent with their role as the primary safety objectives. The fitness surface is smooth and unimodal within each β slice, indicating no abrupt sensitivity to small weight perturbations.

Figure 8a (middle) shows, for the same grid, whether the baseline configuration ($\alpha$=0.35, $\beta$=0.30, $\gamma$=0.35, marked ★) remains the optimal choice (blue) or is displaced by a different configuration (orange). The baseline sits firmly within the blue region across the practically relevant weight range. It is only displaced in the orange cells, where weights are strongly imbalanced — specifically when the speed score weight is pushed to an extreme high (upper-left region of each panel) or the detection weight is dominant with very low speed weight (lower-right region), representing severe over-prioritisation of a single objective.

Figure 8a (bottom) quantifies this further. Across all 90 weight combinations, only four distinct configurations ever emerged as optimal. The baseline configuration was the most dominant, winning in 50% of cases (45 of 90), while the remaining three alternatives each won under increasingly extreme weight assignments. The histogram shows that the baseline configuration's composite score ranged from 0.80 to 0.93 with a mean of 0.87, remaining competitive across the entire weight space. Taken together, these results confirm that the chosen weights are not a critical hyperparameter and that the qualitative conclusions are robust to reasonable variations in the weighting scheme.

### 3.2.4 Sensor Configurations and Coverage

Figure 9 shows plan-view visualizations of the parking garage, comparing the sensor placements obtained by the four methods. In each subfigure, a consistent symbology is used: sensors are represented by black dots, their detection zones by black circles with an assumed detection radius of 7.85 m, and shaded regions in varying colors indicate the number of sensors covering each area using color-coded coverage intensity. Figure 10 quantifies the spatial coverage characteristics visualized in Figure 9 (red: uncovered; orange: single coverage; yellow: double coverage; light green: triple coverage; dark green: quadruple + coverage).

**(a) Random Baseline Configuration**

The random baseline places 15 sensors subject only to a minimum spacing constraint (≥5 m apart), with positions otherwise assigned arbitrarily. Notable characteristics include clustering in the southwest quadrant—four sensors within one detection radius—resulting in over-covered areas, while the northeast and upper right corner remains poorly protected, creating large blind spots, which shows irregular coverage (23.3% total uncovered) with this inefficient clustering. Additionally, one sensor is positioned less than 1 m from walls, violating practical installation guidelines. This configuration serves as a lower-bound performance reference, illustrating the inadequacy of ad-hoc placement strategies.

**(b) Uniform Grid Configuration**

The uniform grid arranges 15 sensors in a regular 5×3 array, with 10 m longitudinal spacing and 10 m lateral spacing. All sensors are located at practical ceiling-mount heights (z = 2.75 m). This layout provides systematic coverage, eliminating blind spots entirely. The statistic histogram displays systematic coverage with double coverage across 53.5% of facility and reasonable triple+ coverage (5%) reflect systematic geometric spacing. Representing current industry practice, this baseline is simple to implement and reasonably effective, yet it lacks optimization for specific leak scenarios.

**(c) GA-Optimized Configuration**

The GA-optimized layout demonstrates strategic improvements: sensors are positioned to balance redundancy and coverage extension effectively. Leveraging ventilation patterns, sensors are preferentially placed downstream of potential leak sources. Compared to the random configuration, blind spots (0.12%) are substantially reduced, and over-covered zones (0%) are minimized while redundancy is enhanced. The efficient 49.6% double coverage and 3.53% triple coverage allocation demonstrate multi-objective optimization balancing coverage extension and strategic redundancy allocation. Despite emerging from an evolutionary optimization process, the resulting configuration exhibits a clear, purposeful structure: sensors form an irregular pattern that directly responds to the facility's specific leak distribution and fluid dynamics.

**(d) DeepSets-Predicted Configuration**

The sensor configuration identified by the DeepSets closely resembles the GA-optimized layout, which is expected given that the surrogate was trained exclusively on configurations evaluated by the GA's true CFD fitness function. The GA population history serves as the ground truth for surrogate training, so the learned fitness landscape inherently reflects the same objective that guided the evolutionary search, biasing the gradient optimization toward solutions in regions the GA found most promising. The spatial similarity between the two configurations is therefore a direct consequence of this training regime rather than coincidental convergence. The moderate surrogate $R^2$ of 0.80 introduces a small degree of prediction error, which manifests as minor positional deviations of individual sensors relative to their GA counterparts.

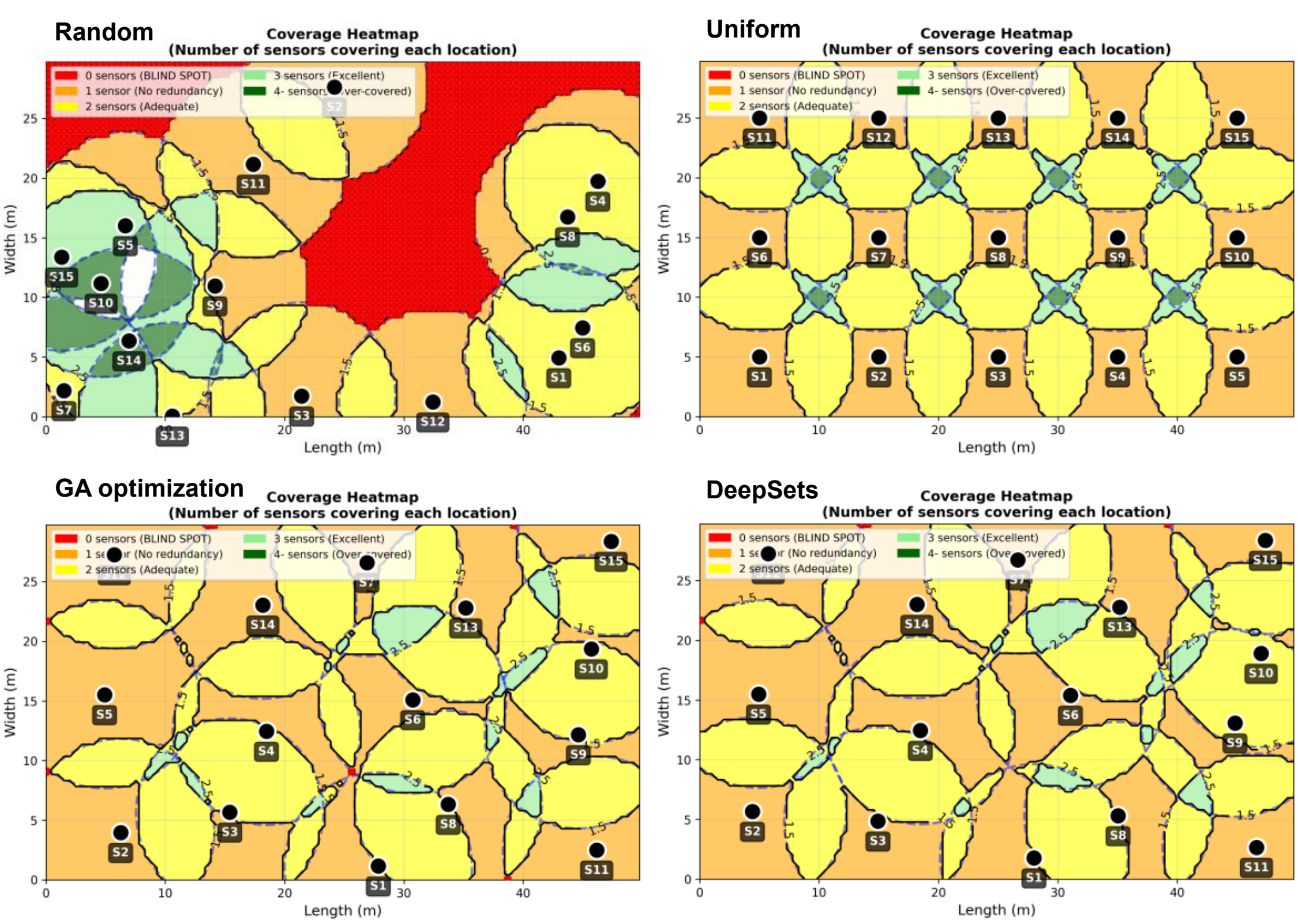

Figure 9 Sensor placement (2D Top View) and coverage heatmaps comparison

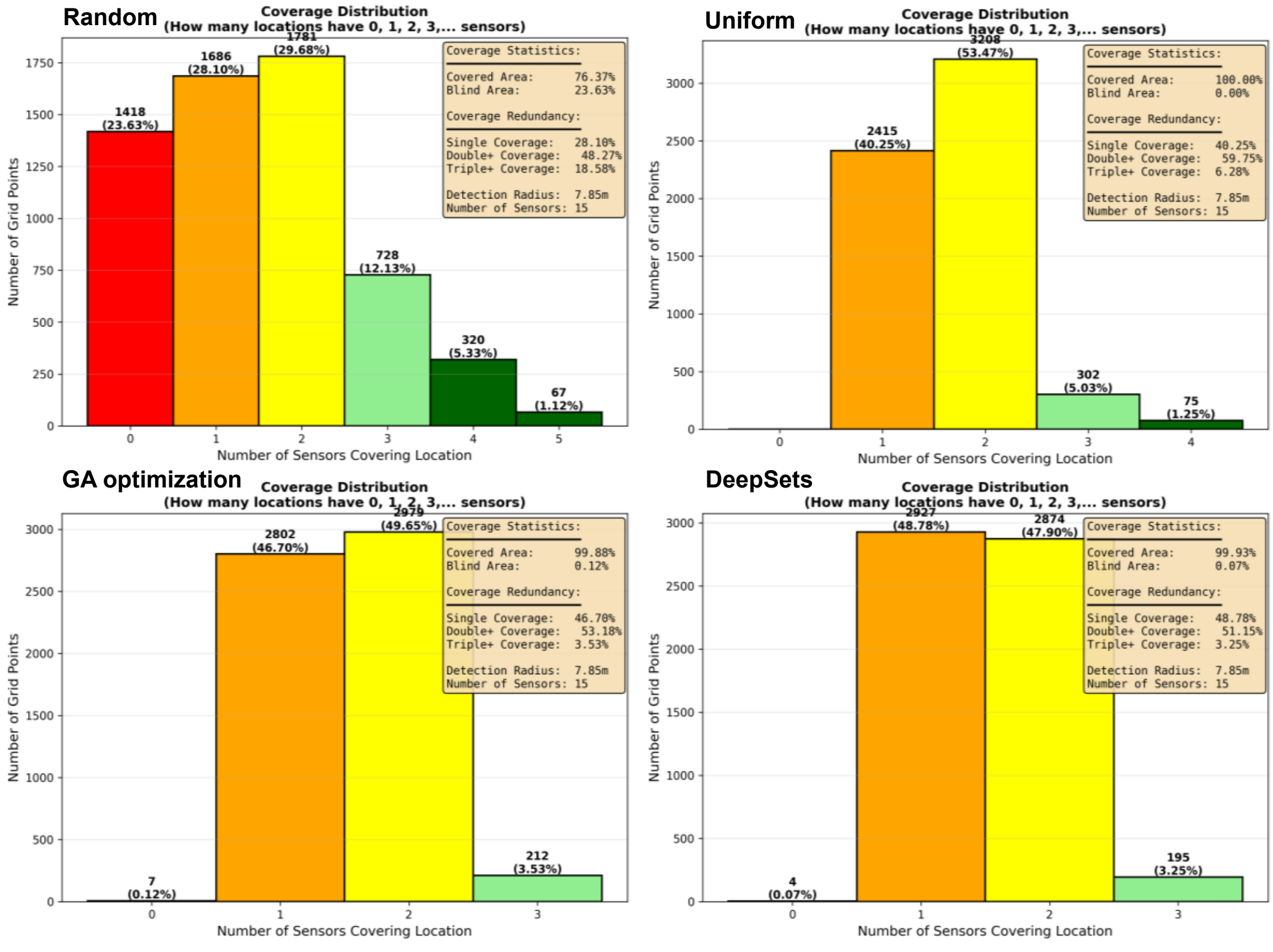

Figure 10 Coverage distribution comparison

### 3.2.5 Detection Rate and Timing

Figure 11 presents cumulative detection rate vs. time curves for all four methods, showing the percentage of 180 leak scenarios where at least one sensor detected hydrogen concentrations exceeding the 0.1 vol. % threshold within 60 seconds since leak initiation. The histograms of each method in Figure 12 show detection time frequency distribution with ~3-second bins (x-axis: 0-60s; y-axis: number of scenarios). Table 3 summarizes statistical distributions of detection performance across the 180 scenarios, using the critical metrics: detection rate and timing. These metrics directly measure safety effectiveness—undetected leaks represent system failures. The results of stratified detection rate table by leak rate are provided in the Supplementary Material.

Table 3 Detection performance comparison

| Method | Detection Rate in 60 s | Mean/median Detection Time (s) | Best/worst Detection Time (s) | 80% Detection Time (s) | 95% Detection Time (s) |
|---|---|---|---|---|---|
| Random | 86.7% (156/180) | 9.37/6 | 2/58 | 19.5 | ~ |
| Uniform | 92.2% (166/180) | 9.9/7 | 2/59 | 15 | ~ |
| GA | 96.1% (173/180) | 7/4 | 1/60 | 6.1 | 57 |

| DeepSets | 95.6% (172/180) | 7.36/4 | 1/60 | 7.4 | 58 |
|---|---|---|---|---|---|

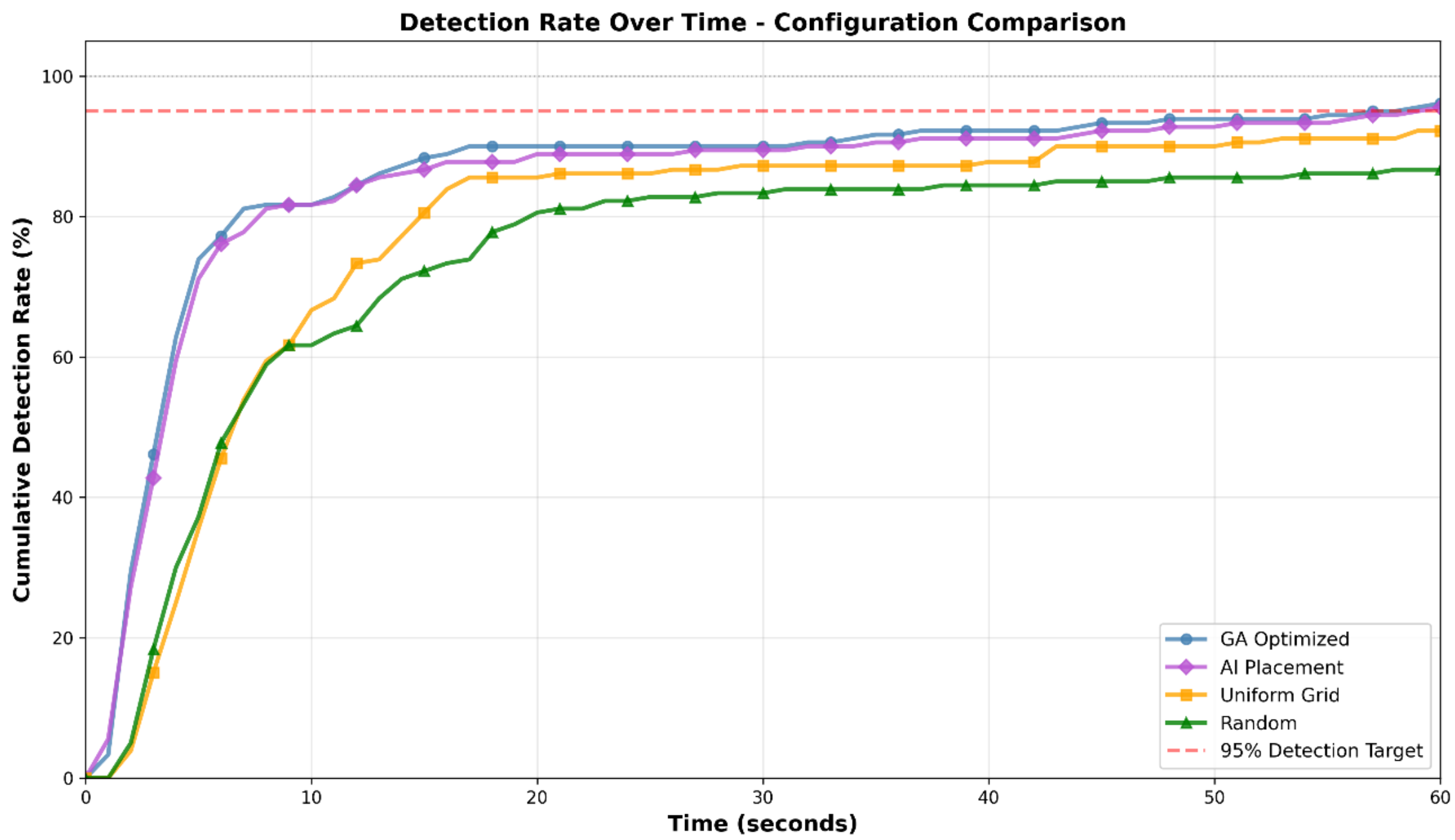

Figure 11 Detection rate over time comparison

- **Random baseline**: Failed to detect 24 of 180 scenarios, primarily low-rate leaks (1 g/s) under high ventilation (ACH = 10 $h^{-1}$), where dilution reduced ceiling concentrations below detection thresholds. The resulting 13.3% failure rate represents an unacceptable risk for safety-critical applications, lacking early-warning capability for many scenarios. The detection rate curve (green line in Figure 11) never reaches 95%, plateauing at 86.7% as 24 scenarios remain undetected even at the 60s timeout.

- **Uniform grid**: Improved coverage reduced failures to 14 scenarios (7.8%), all involving low leak rates under high ventilation. While systematic spacing eliminated large blind spots, the layout lacked strategic positioning for flow dynamics of scenarios. The detection rate curve (orange line in Figure 11) shows characteristic two-phase behavior—rapid initial detection (85.3% by t=17.4s) followed by plateau approaching 92.2% asymptote. It also fails to achieve the 95% detection target within 60 seconds.

- **GA-optimized**: Only 7 missed detections out of 180 scenarios—halving the failure rate compared to the uniform grid. Failures occurred exclusively in extreme edge cases, where the combination of minimal hydrogen mass, maximum dilution, and peripheral location resulted in concentrations below 0.1% vol. at all sensor positions. The detection rate curve (blue line in Figure 11) exhibits sustained detection accumulation through t=6.1 s up to 80%, with 95% of scenarios detected within 57 s.

- **DeepSets-Predicted**: Detected 172 of 180 scenarios (95.6%). The single additional missed detection occurs exclusively at the lowest leak rate (1 g/s), where temporal dispersion dynamics are most critical. DeepSets pipeline achieves detection rate and times comparable to the GA, indicating that the surrogate captures early transient fitness contributions with sufficient fidelity to preserve rapid-detection behavior.

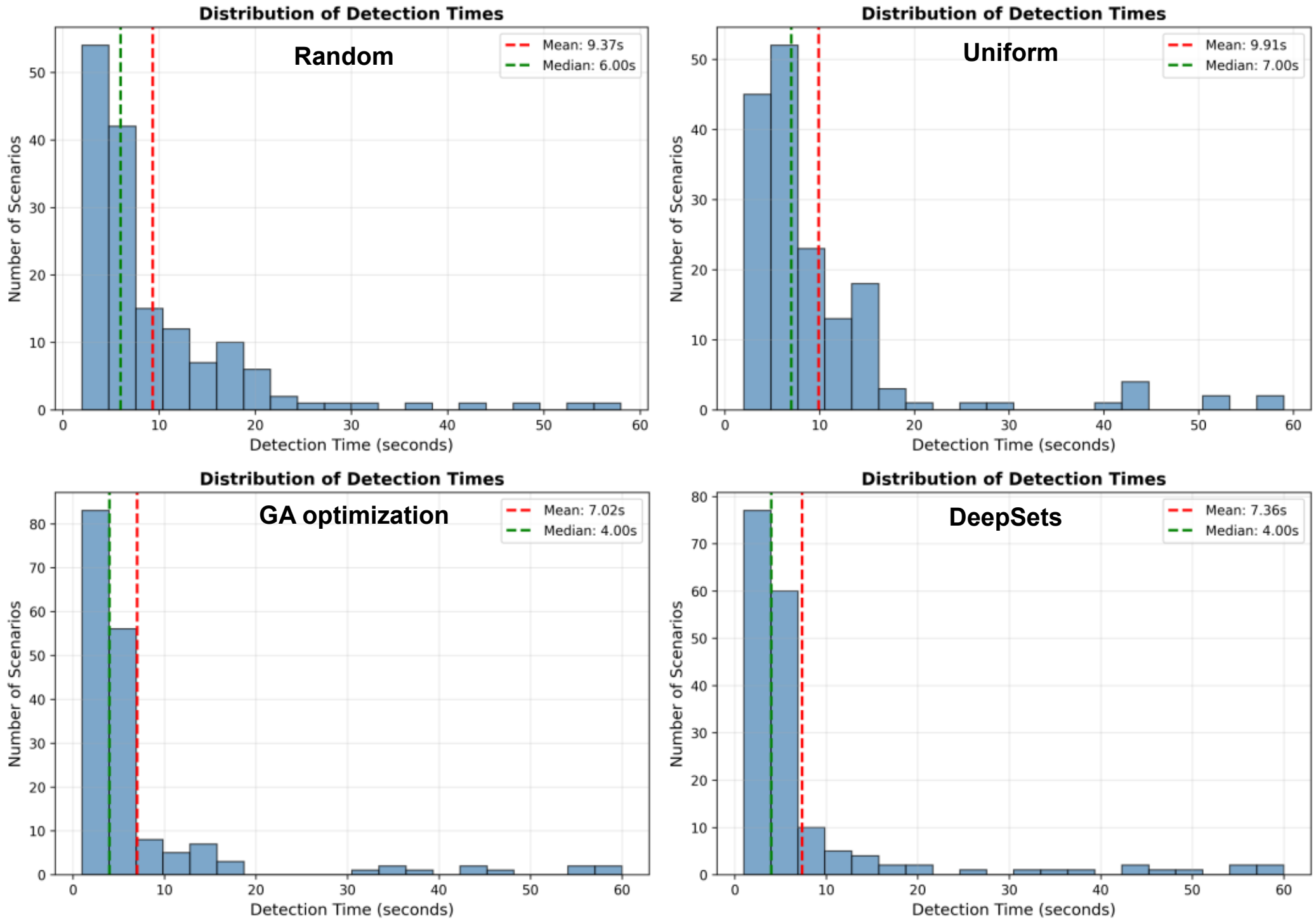

Figure 12 Distribution of detection times

# 4 Discussions

## 4.1 GA Outperforms Baselines

The superior performance of the genetic algorithm compared to baselines arises from its ability to systematically exploit the spatial and temporal patterns revealed by the CFD simulations. While the uniform grid treats all locations equivalently, the GA leverages a multi-objective fitness function to identify sensor positions that simultaneously maximize detection probability, spatial coverage, and response time.

In particular, the detection component enhances performance for low leak-rate scenarios and reduces detection time by prioritizing early-response configurations. This data-driven optimization enables the algorithm to identify effective sensor placements that are not obvious from geometric intuition alone. The coverage component further eliminates blind spots compared to random placement, especially in regions where hydrogen accumulation deviates from uniform dispersion due to buoyancy and ventilation effects.

As a result, the GA discovers non-intuitive yet physically meaningful configurations—for example, placing sensors slightly offset from geometric centroids to better intercept buoyancy-driven plumes and ventilation-induced flow paths. These placements would be difficult to derive using rule-based or heuristic methods. Overall, the GA translates CFD-informed dispersion characteristics into actionable sensor layouts, leading to measurable improvements in detection rate, timing, and robustness.

## 4.2 DeepSets Benefits and Gap

Table 4 indicates the performance comparison of GA and AI DeepSets sensor placement. The final AI configuration achieves a composite fitness of 0.8165 and a detection rate of 0.9556, matching the GA exactly on detection rate and closing to within 0.01 on composite fitness. Coverage score reaches 0.8991versus 0.9015 for the GA, and the blind spot is 0.07% compared to the GA's 0.12%. These results are achieved using 94% fewer true CFD evaluations (800 vs 11,307) and reducing wall-clock time from 144.6 minutes to under 1 minute. The AI configuration achieves 77.8% detection at 1 g/s versus the GA's 80.6%, a difference of 2.8% (1 in 36 of 1g/s scenarios) that reflects the lower composite fitness driven by the penalty and speed score term. Detection is perfect (100%) for all higher leak rates (≥30 g/s). Mean detection time is 7.36 s versus 7.02 s for the GA, a marginal difference of 0.34 s. The remaining gaps — speed score (0.731 vs 0.742) and dual-sensor coverage (51.2% vs 53.2%) — are consistent with the surrogate approximation error and do not represent a structural limitation of the approach.

Table 4 Performance comparison of GA and AI DeepSets sensor placement

| Metric | GA | AI DeepSets |
|---|---|---|
| **Composite Fitness** | 0.8260 | 0.8165 |
| **Detection Rate** | 0.9611 | 0.9556 |
| **Speed Score** | 0.7420 | 0.7308 |
| **Coverage Score** | 0.9015 | 0.8991 |
| **Penalty** | 0.0406 | 0.0435 |
| **Blind Spot (%)** | 0.12 | 0.07 |
| **≥ 2-Sensor Cov.** | 53.18 | 51.15 |
| **1 g/s Detection (%)** | 80.6 | 77.8 |
| **True CFD Evals** | 11,307 | 800 |
| **Runtime (min)** | 142.9 (-20 data loading) | 0.5 |

DeepSets provides a fundamentally different advantage over the GA by enabling fast, gradient-based optimization in a continuous design space. Unlike the GA, which relies on iterative population-based search with many expensive CFD evaluations, DeepSets learns an approximate fitness landscape from prior data and can therefore identify high-quality sensor configurations with orders-of-magnitude lower computational cost. This makes it particularly valuable for rapid design iterations and scalability to larger or more complex facilities. However, the remaining performance gap relative to the GA stems from surrogate approximation error, especially in penalty and speed score term. Despite this, DeepSets achieves near-equivalent performance, demonstrating that learned surrogate models can effectively replace most expensive simulations while preserving solution quality.

## 4.3 Practical Implications

The comparison between GA and DeepSets reveals a clear trade-off between solution optimality and computational efficiency, which directly informs practical deployment strategies. GA provides the highest-quality sensor configurations, achieving marginally better detection performance and redundancy, but at the cost of substantial computational effort. In contrast, DeepSets delivers near-equivalent performance (within ~1% fitness difference) with over an order-of-magnitude reduction in computational cost, making it more suitable for rapid design iteration and scalability to multiple facilities.

From an economic perspective, both optimized approaches reduce sensor requirements compared to uniform layouts (15 GA-optimized vs. 18 uniform sensors for comparable detection performance), corresponding to approximately 17% savings in hardware and installation. In this context, the GA is best suited for high-stakes, one-time design optimization, while DeepSets enables efficient reuse and adaptation of optimized configurations across similar environments.

For implementation, the CFD-informed optimization results suggest that sensor placement should prioritize ceiling-mounted locations aligned with dominant flow and accumulation patterns rather than uniform spacing. Integration with ventilation control systems further enhances safety performance, as early detection enables timely activation of high-rate ventilation, significantly reducing flammable cloud formation. The surrogate-based DeepSets framework additionally supports real-time monitoring and digital twin applications, where rapid evaluation is essential.

# 5 Conclusion

This study presents a computational framework for optimizing hydrogen leak detection sensor placement in enclosed environments by integrating CFD simulations, genetic algorithm optimization, and DeepSets neural surrogate. A CFD database of 180 scenarios, spanning multiple leak positions, rates, and ventilation conditions, enables robust and systematic evaluation.

The GA achieves high detection performance (96.1% within 60 s) while minimizing blind areas (0.12%), outperforming conventional uniform layouts by approximately 5% in composite fitness. In parallel, the DeepSets surrogate provides a computationally efficient alternative, reproducing near-optimal configurations with a fitness gap below 0.01. This is achieved with an 89% reduction in CFD evaluations, and a two-order-of-magnitude decrease in runtime, while maintaining comparable detection performance and spatial coverage.

The complementary roles of GA and DeepSets enable both high-quality one-time optimization and rapid re-optimization for changing layouts, supporting practical deployment across multiple facilities.

These results demonstrate that CFD-informed optimization significantly improves both detection effectiveness and resource efficiency. The study further contributes: (1) a large-scale CFD scenario database for hydrogen dispersion; (2) a systematic comparison of optimization strategies (GA, DeepSets, uniform, random); (3) a multi-objective fitness formulation incorporating detection probability, timing, and coverage; and (4) a surrogate-assisted optimization framework suitable for rapid design iteration.

Future work should validate the proposed approach in a real-world or experimental facility to determine the optimal sensor configuration and the minimum number of sensors required for reliable hydrogen leak detection. The methodology should also be extended to more complex geometries and dynamic operating conditions, including industrial facilities and transport infrastructures. The proposed framework provides a generalizable foundation for the design and operation of advanced hydrogen safety systems. Integration with digital twin frameworks further enables a shift from reactive leak detection to proactive risk assessment and decision support.

## Acknowledgement

The authors gratefully acknowledge the fruitful discussions and valuable comments provided by William Buttner, Munjal Purnkant Shah, David Peaslee, Logan Martens from the National Laboratory of the Rockies (NLR). Their insightful feedback and technical suggestions have significantly improved the quality of this work.

# Acknowledgement

# 1 DeepSets Neural Surrogate

## 1.1 Trust-Region Gradient Optimizations

### 1.1.1 Motivation for the Trust Region

With a trained surrogate in hand, sensor placement is framed as a continuous optimization problem: maximize the surrogate-predicted composite fitness subject to facility bounds and minimum sensor spacing constraints. Gradient-based optimization via L-BFGS-B is applied, using finite-difference gradients computed through the surrogate.

A naïve application of gradient optimization to a neural surrogate, however, tends to fail in a characteristic way: the optimizer confidently ascends toward surrogate maxima in regions of the input space that are sparsely represented in the training data. Because the surrogate has not been trained in such configurations, its predictions in those regions are unreliable and systematically overestimated. In preliminary experiments without regularization, this produced a surrogate-to-CFD fitness gap of −0.11, rendering the optimized configuration effectively useless when validated against true physics.

### 1.1.2 Trust-Region Penalty

To address this, a quadratic trust-region penalty is added to the objective function that penalizes solutions straying more than a radius r from the restart's seed configuration:

$$\mathrm{P_{TR}}(x, x_0) = \lambda \cdot max\left(0, \left\|x_{xy} - x_{0,xy}\right\|^2 - r^2\right) \qquad \text{Equation S11}$$

where $x_{xy}$ denotes the (x, y) sensor coordinates (z is fixed at ceiling height), $x_0$ is the seed configuration, λ = 0.5 is the penalty weight, and r = 5 m is the trust radius. The trust region keeps each optimization restart within the region of the input space covered by its seed — a configuration whose fitness has been verified by true CFD evaluation — thereby preventing excursions into unreliable surrogate territory.

Seeds for the restarts are selected from the top-fitness labelled configurations in the training set, ensuring each restart begins at a known-good location. With trust radius r = 5 m, the surrogate-to-CFD fitness gap reduces to near zero (Δ = +0.017 in the current run), and the validated composite fitness reaches 0.816–0.820 across runs depending on the coverage strategy described in Sections 3.5 and 3.6.

### 1.1.3 Wall-Margin Enforcement

Physical installation constraints require sensors to be placed at least 1 m from every wall. This constraint is enforced at three levels of the pipeline. First, hard bounds in the L-BFGS-B optimizer restrict the feasible region to [WALL_MARGIN, L − WALL_MARGIN] × [WALL_MARGIN, W − WALL_MARGIN]. Second, a quadratic penalty term in the objective function provides gradient information pushing solutions away from the boundary. Third, and critically, every seed configuration is clipped to the valid bounds before it is used as a trust-region anchor.

The third step resolves a subtle conflict that arose in practice: GA population members sometimes placed sensors very close to walls. When such a configuration was used directly as a trust-region anchor, the trust-region penalty gradient actively pulled the optimized solution back toward the wall, fighting the hard bounds and causing sensors to be pinned at exactly WALL_MARGIN rather than moving freely into the interior. Clipping the anchor first ensures the trust-region gradient always points away from walls.

## 1.2 Random Restart Augmentation

The GA population history is not uniformly distributed across the facility: evolutionary pressure concentrates configurations in the parts of the design space the GA found most promising, often leaving other regions sparsely sampled. When all 20 optimization restarts are seeded exclusively from GA history, the trust-region bubbles collectively cover only the well-explored regions, potentially leaving gaps in facility coverage.

This effect was observed empirically: with GA-only seeds and a 5 m trust radius, several restarts were anchored near the right boundary of the facility and explored only x ≈ 44–49 m, producing a visible blind spot in the central region of the AI-optimized layout.

To address this, 5 of the 20 restarts are seeded from uniformly random configurations generated by the wall-safe random initializer rather than from GA history. These random seeds are distributed across the full facility interior, and their trust-region bubbles explore regions the GA never concentrated on. The remaining 15 restarts continue to use GA-seeded starts, preserving the benefit of beginning near known high-fitness configurations.

This hybrid seeding strategy — 15 GA-seeded restarts plus 5 random restarts — reduced the blind spot from 1.13% to 0.85% of the facility area and improved the composite fitness from 0.879 to 0.882 in the development phase. However, sensors continued to cluster in GA-explored subregions of the facility because the trust-region gradient has no incentive to explore uncovered areas elsewhere.

## 1.3 Coverage Grid Bonus

Even with hybrid seeding, restarts anchored in GA-explored regions have trust-region bubbles centered away from under-sampled areas of the facility. The trust-region penalty gradient provides no incentive to cover regions that lie outside a restart's 5 m reach, resulting in a residual blind spot even when the global coverage score appears high.

To address this, a coverage grid bonus is added directly to the optimization objective. The facility floor is discretized into a 25×15 grid of cells. For each cell center c, a soft coverage score is computed as a sum of Gaussian proximity kernels across all sensors, clipped to [0, 1]:

$$\mathrm{cov}(c) = \min\left(1, \sum_{k} \exp\left(-\|c - s_k\|^2 / {det_r}^2\right)\right) \qquad \text{Equation S12}$$

The mean coverage across all grid cells is added as a bonus term (scaled by $w_{cov} = 2.5$) to the objective function being maximized. Only unsaturated cells (cov < 1) contribute gradient, so sensors already in well-covered regions experience no force. This provides a strong, spatially targeted gradient: a blind spot at any grid cell pulls the nearest sensor directly toward it. Unlike a 1/d diversity penalty — which produces negligible gradients (~$4\times10^{-6}$) at typical 10–15 m inter-sensor separations — the coverage grid gradient is of order $10^{-2}$–$10^{-3}$, sufficient to meaningfully reposition sensors within their trust radius.

With $w_{cov} = 2.5$, the blind spot in the validated run is 0.07% and the composite fitness reaches 0.817. The small residual blind spot approaches the practical floor: sensors can only move 5 m from their trust-region anchors, so facility corners unreachable from any GA seed cannot be covered regardless of the coverage gradient weight.

# 2 Stratified detection rate table by leak rate

Table S1 presents the detection rate stratified by leak rate for all four sensor placement methods across the 180 CFD scenarios (36 scenarios per leak rate: 12 leak positions × 3 ventilation levels). At all leak rates of 30 g/s and above, the GA-optimized and DeepSets configurations achieve 100% detection, confirming that the optimized layouts provide complete coverage under moderate-to-high release conditions. Detection failures are confined exclusively to the 1 g/s scenario, where hydrogen mass flow is minimal and high ventilation (ACH = 10 $h^{-1}$) dilutes concentrations below the 0.1 vol.% threshold at all sensor positions. Under these most challenging conditions, the GA and DeepSets achieve 80.6% detection (29/36 scenarios), compared to 61.1% for the uniform grid and only 44.4% for the random baseline—representing improvements of 32% and 81% respectively over the two conventional layouts. The random baseline also shows minor failures (97.2%) even at higher leak rates (30–150 g/s), reflecting the impact of its suboptimal sensor clustering. These results confirm that optimization-based methods provide the most robust safety margins across the full range of credible leak scenarios, with their advantage being most pronounced precisely in the low-leak, high-ventilation conditions that are most difficult to detect and most relevant to early-warning performance.

**Table S1.** Detection rate stratified by leak rate for all four sensor placement methods (36 scenarios per leak rate).

| Leak rate (g/s) | Random (%) | Uniform (%) | GA (%) | DeepSets (%) |
|---|---|---|---|---|
| **1** | 44.44 (16/36) | 61.11 (22/36) | 80.56 (29/36) | 80.56 (29/36) |
| **30** | 97.22 (35/36) | 100.00 (36/36) | 100.00 (36/36) | 100.00 (36/36) |
| **50** | 97.22 (35/36) | 100.00 (36/36) | 100.00 (36/36) | 100.00 (36/36) |
| **100** | 97.22 (35/36) | 100.00 (36/36) | 100.00 (36/36) | 100.00 (36/36) |
| **150** | 97.22 (35/36) | 100.00 (36/36) | 100.00 (36/36) | 100.00 (36/36) |